\documentclass[letterpaper, 10 pt, conference]{ieeeconf}  

\IEEEoverridecommandlockouts                              
\usepackage{amsmath} 
\usepackage{amssymb}  
\usepackage{graphicx}

\usepackage[subrefformat=parens,labelformat=parens]{subcaption}
\usepackage[export]{adjustbox} 

\usepackage{cleveref}
\crefname{figure}{Fig.}{Figs.}

\usepackage{booktabs,tabularx}
\usepackage{float}
\usepackage{bm}

\newcolumntype{Y}{>{\centering\arraybackslash}X}

\newcommand{\point}[0]{\mathbf{p}}

\newcommand{\param}{{\Theta}} 
\newcommand{\inp}{{\lambda}} 
\newcommand{\dhp}{{\omega}} 
\newcommand{\sdof}{{\tau}} 

\newcommand{\Tf}[2]{\mathbf{T}^{#1:#2}}
\newcommand{\Tpf}[3]{\mathbf{T}^{#2:#3}_{#1}}
\newcommand{\Tpaf}[4]{\mathbf{T}^{#3:#4}_{#1,#2}}

\newcommand{\proj}[0]{\Psi}

\newcommand{\card}[1]{|#1|}
\newcommand{\fr}[1]{\mathcal{F}_{#1}}

\newcommand{\dycam}{d}
\newcommand{\stcam}{s}
\newcommand{\target}{t}
\newcommand{\ee}{e}

\title{\LARGE \bf
Encoderless Gimbal Calibration of Dynamic Multi-Camera Clusters
}

\author{Christopher L. Choi$^{1}$, Jason Rebello$^{2}$, Leonid Koppel$^{3}$, Pranav Ganti$^{4}$, Arun Das$^{5}$, and Steven L. Waslander$^{6}$
\thanks{$^{1}$MASc Candidate, 
		Mechanical and Mechatronics Engineering,
        University of Waterloo,
        {\tt\small c33choi@uwaterloo.ca}}%
\thanks{$^{2}$PhD Candidate, 
		Mechanical and Mechatronics Engineering,
        University of Waterloo,
        {\tt\small jrebello@uwaterloo.ca}}%
\thanks{$^{3}$MASc Candidate, 
		Mechanical and Mechatronics Engineering,
        University of Waterloo,
        {\tt\small lkoppel@uwaterloo.ca}}%
\thanks{$^{4}$MASc Candidate, 
		Mechanical and Mechatronics Engineering,
        University of Waterloo,
        {\tt\small pganti@uwaterloo.ca}}%
\thanks{$^{5}$PhD Candidate, 
		Mechanical and Mechatronics Engineering,
        University of Waterloo,
        {\tt\small adas@uwaterloo.ca}}%
\thanks{$^{6}$Associate Professor, 
		Mechanical and Mechatronics Engineering,
        University of Waterloo,
        {\tt\small stevenw@uwaterloo.ca}}%
}

\begin{document}

\maketitle
\thispagestyle{empty}
\pagestyle{empty}

\begin{abstract}
Dynamic Camera Clusters (DCCs) are multi-camera systems where one or more cameras are mounted on actuated mechanisms such as a gimbal. Existing methods for DCC calibration rely on joint angle measurements to resolve the time-varying transformation between the dynamic and static camera. This information is usually provided by motor encoders, however, joint angle measurements are not always readily available on off-the-shelf mechanisms. In this paper, we present an encoderless approach for DCC calibration which simultaneously estimates the kinematic parameters of the transformation chain as well as the unknown joint angles. We also demonstrate the integration of an encoderless gimbal mechanism with a state-of-the art VIO algorithm, and show the extensions required in order to perform simultaneous online estimation of the joint angles and vehicle localization state. The proposed calibration approach is validated both in simulation and on a physical DCC composed of a 2-DOF gimbal mounted on a UAV. Finally, we show the experimental results of the calibrated mechanism integrated into the OKVIS VIO package, and demonstrate successful online joint angle estimation while maintaining localization accuracy that is comparable to a standard static multi-camera configuration.
\end{abstract}

\section{Introduction}
\label{sec:introduction}

	Multi-camera clusters (MCCs) have emerged as an effective camera configuration for robust vision-based localization and mapping algorithms~\cite{tribou2015multi, heng2015self}. In contrast to a standard stereo setup, which exploits the overlap between two forward facing cameras, MCCs use multiple cameras in a wider range of configurations. The increased field-of-view (FOV) available to MCCs facilitates feature tracking over longer durations and larger viewpoint changes in comparison to monocular or stereo configurations, which can lead to improved pose estimates for robotic applications.
    
	 MCCs can be set up in two possible configurations: static camera clusters (SCCs) or dynamic camera clusters (DCCs). SCCs have rigidly mounted camera configurations with static extrinsic parameters, while DCCs incorporate a camera mounted to an actuated mechanism, such as a gimbal, allowing the cameras to reconfigure their orientation independent of each other and irrespective of robot motion. Although SCCs have provided promising results thus far, the features observed by this configuration are directly coupled to the motion of the camera cluster. A trajectory where the cameras observe limited features or occlusions can drastically degrade pose estimates, resulting in efforts to solve this problem by placing the entire SCC on a swiveling rig~\cite{manderson2016texture, frintrop2008attentional}. In contrast to an SCC, DCCs allow for a more general camera configuration with the ability to actively select image viewpoints. This dynamic viewpoint manipulation enhances the tracking ability of interesting and informative features, which can lead to improved localization estimates. The drawback of active DCC viewpoint selection, however, is that the time-varying camera extrinsics must be resolved with sufficient accuracy to perform visual odometry or SLAM without negatively impacting estimation performance.

	The authors' previous work outlines the calibration of time-varying extrinsic parameters for various DCC configurations, such as one static and one actuated camera as well as two actuated cameras~\cite{das2016calibration}. The DCC calibration was also extended to determine optimal gimbal configurations which locally minimize parameter uncertainty using next-best-view~\cite{rebello2017autonomous}. Although the proposed approaches effectively perform DCC calibration, both methods rely on having encoder feedback to extract the position of the gimbal motor, and also require knowledge of the joint angles to determine the kinematic chain between the static and actuated cameras.
    
    In practice, it is not always feasible to have access to the joint angle information of the actuated mechanism, and in some cases, the integration of an encoder may not be desirable due to weight or cost limitations. Many off-the-shelf gimbals found on multirotor Unmanned Aerial Vehicles (UAVs) either do not provide joint angle information, or provide imprecise joint angle measurements which cannot be used to obtain accurate DCC calibration. Manually adding encoders is not always possible, as most commercial gimbals come as integrated units with custom hardware and software which cannot be easily modified. The lack of joint angle feedback also introduces difficulties while performing VIO or SLAM algorithms, as accurate joint angle information is beneficial to resolving the time-varying extrinsics during robot motion. 
    
	In this work, we present two main contributions. First, we develop an \emph{encoderless} DCC calibration approach which eliminates the requirement of joint angle measurements from the actuated mechanism. Our approach estimates the DCC calibration parameters in conjunction with the joint angles of the actuated mechanism for each configuration of the collected measurement set. By eliminating the need for joint angle information, this calibration can be performed on any actuated mechanism, allowing for visual SLAM algorithms to be run on any off-the-shelf DCC. Second, we extend the Visual Inertial Odometry (VIO) algorithm developed in~\cite{leutenegger2015keyframe} and incorporate the calibrated DCC parameters in order to simultaneously estimate the gimbal joint angles and the VIO localization state. We obtain experimental results using a gimbal based DCC mounted on a custom quadrotor platform, and evaluate the relative performance of SCCs and DCCs in a large outdoor environment with a variety of static and gimballed camera configurations. We show that our overall pose estimate using a DCC is comparable to those obtained using an SCC configuration without significant change in performance.

\section{Related Works}
\label{sec:related_works}


DCC calibration requires the estimation of parameters which define the extrinsic transformation between a static camera and a camera mounted to an actuated mechanism, such as a robot manipulator or gimbal. Since DCC calibration involves characterizing the kinematic parameters of the actuated mechanism, it is closely related to other calibration problems such as hand-in-eye~\cite{horaud1995hand, remy1997hand}, hand-to-eye~\cite{chen1993new}, head-to-eye~\cite{kim2010robot} and kinematic calibration~\cite{pradeep2014calibrating}. These methods, however, do not solve the problem of a camera to camera calibration through an actuated mechanism, which is essential to resolve measurements from multiple cameras when performing visual odometry and SLAM. The calibration of the time-varying extrinsics for a DCC was achieved by Das and Waslander~\cite{das2016calibration}; this approach uses the Denavit-Hartenberg (DH) parameters~\cite{hartenberg1964kinematic} to model the forward kinematics of the actuated link as a serial manipulator, and then estimates the calibration parameters using overlapping viewpoints of a fiducial target. The work in~\cite{das2016calibration} was later extended to perform fully automatic viewpoint selection and calibration using a locally optimal next-best-view approach~\cite{rebello2017autonomous}. Although the existing DCC approaches provide good calibration results, the proposed methods require accurate measurements of the mechanism's joint angles. In many practical cases, accurate joint angle measurements are not accessible, thus an \emph{encoderless} joint angle estimation approach is required.   

Existing encoderless approaches found in literature mainly attempt to perform joint motor control without the use of velocity and position sensors mounted on the motor shaft. Kawamura et al.~\cite{kawamura2014encoderless} propose augmenting the motor control signal of a manipulator joint with a high frequency component, then measuring the changes in the motor's back-EMF. This approach estimates the joint angle positions through the use of a known dynamic model for the motor and external measurements of the end-effector position. Baik et al.~\cite{baik2004position} suggest a method to perform motor angle estimation of a Switched Reluctance Motor (SRM) using a neural network. Using a temporary external reference system, the system is trained to estimate the motor angle using the measured motor flux and current as inputs to the network. Since a typical off-the-shelf gimbal system does not provide measurements of the motor back-EMF and motor flux, these existing approaches are not suitable to apply to our joint angle estimation problem. The work of Kormushev et al.~\cite{kormushev2015encoderless} demonstrates a learning algorithm capable of positioning the end-effector of a manipulator without the knowledge of joint angles or the robot's kinematic parameters. By generating random control signals and observing the end-effector, they show that the local kinodynamics of the manipulator can be approximated and the required control signal for end effector positioning can be estimated. Although their approach is able to perform accurate positioning of the end-effector without joint angle measurements, the proposed learning algorithm requires occasional exploratory motions of the end-effector in order to calculate the local kinodynamics. Application of the learning algorithm to our gimbal joint angle estimation problem is impractical, as the required exploratory motions would interrupt the calibration or localization process.


Joint angle estimation can also be thought of as an \emph{online} calibration problem; once the DCC is calibrated, the mechanism's joint angles still need to be estimated simultaneously with the vehicle's localization state as the mechanism changes configurations. Existing online calibration approaches generally augment the state estimate with calibration parameters in order to resolve or refine extrinsic sensor relationships. For example, the camera to IMU extrinsic calibration~\cite{li20133, jia2013online, weiss2012real} and IMU to body frame calibration~\cite{joukov2017generalized}, can be successfully refined during online operation, provided the IMU experiences sufficient excitation. K\"ummerle et al.~\cite{kummerle2012simultaneous} present an approach that simultaneously performs calibration, mapping, and localization using a graph-based nonlinear least squares framework. Although accurate calibration performance was achieved, the mapping formulation includes information from a 3D laser scanner which greatly simplifies scale recovery for the solution. For MCCs performing online calibration, the camera extrinsics are only known up to scale, and the true scale cannot be resolved without additional knowledge of the environment or motion of the camera cluster~\cite{carrera2011slam}. The work most related to ours is by Warren et al.~\cite{warren2013online}, who present an online method to refine the extrinsic calibration of a stereo camera. In their approach, the scale of the solution is maintained by assuming that the baseline distance between the stereo pair will never vary significantly from the original or any subsequent calibration. Conversely, in our online joint angle estimation problem, we do not assume that the extrinsic calibration will be a small perturbation as the DCC may assume any valid configuration within the mechanism's configuration space.  

\emph{Active vision} algorithms~\cite{bajcsy1988active} are a primary application of DCCs, as active approaches perform viewpoint manipulation in order to extract useful information from the environment. Existing active vision approaches either use a monocular configuration or perform viewpoint manipulation of a complete SCC; these approaches do not estimate a time varying extrinsic calibration between cameras~\cite{sharp2001vision,playle2015improving,borowczyk2017autonomous,manderson2016texture}.  Gimballed cameras, which are typically found on UAVs for image stabilization, are strong candidates for active vision tasks. Recently, Playle~\cite{playle2015improving} developed an active viewpoint selection policy for a gimballed camera to improve the localization accuracy of a visual SLAM method. Playle's approach, however, also used a monocular camera configuration and therefore did not require a dynamic camera calibration. Our proposed approach to solve DCC calibration and estimate the time-varying camera extrinsic relationship will allow for viewpoint manipulation algorithms to be applied to any dynamic multi-camera cluster by removing the limitation of fixed camera extrinsic relationships.
\section{Background}
\label{sec:background}

\textbf{General Rigid Body Transformation:} Let a 3D point in coordinate frame $\mathcal{F}_x$ be denoted as $\mathbf{p}^x \in \mathbb{R}^3$. We express a rigid body transformation from frame $\mathcal{F}_a$ to $\mathcal{F}_b$ as $\textbf{T}_{\tau}^{b:a} \in \mathbb{SE}(3)$, where $\textbf{T}_{\tau}^{b:a} : \mathbb{R}^3 \mapsto \mathbb{R}^3$, and $\tau = [r_x, \,r_y, \, r_z, \,t_x, \,t_y, \,t_z]$ is the parameter vector used to construct the transformation. The first three components $r_x, \, r_y, \, r_z \in [0,\,2\pi)$ represent 3-2-1 Euler angle rotations, while $t_x, \, t_y, \, t_z \in \mathbb{R}^3$ represent translation values along the respective axes. Although this work uses Euler angles, the rotations can be represented using other conventions, such as quaternions or $\mathbb{SO}(3)$ rotation matrices.

\textbf{Image Projections of 3D Points:} The projection function $\psi(\mathbf{p}_i^c) : \mathbb{R}^3 \mapsto \mathbb{P}^2$, defined as $\psi(\mathbf{p}_i^c) = [u_i^c \: v_i^c]^T$, maps a point $\textbf{p}_i^c$ in camera frame $\mathcal{F}_c$ to a pixel location on the 2D image plane. Here $u_i$ and $v_i$ are the pixel coordinates of the projected point along the $u$ and $v$ dimensions respectively.
	
\textbf{Denavit-Hartenberg Parameterization:} We use the Denavit-Hartenberg (DH) convention to assign coordinate frames to the links of the actuated mechanism, which is modelled as a serial manipulator with rotational joints. The homogeneous rigid body transformation from one frame to another, $\mathcal{F}_{l-1}$ to $\mathcal{F}_l$, can be described by the DH parameters $[\theta_l, \,d_l, \,a_l, \,\alpha_l]^T$, where $\theta_l, \,\alpha_l \in [0, \,2\pi)$ and $d_l, \,a_l \in \mathbb{R}$. We distinguish between $\theta_l$ and the rest of the DH parameters by defining $\omega_l = [\,d_l, \,a_l, \,\alpha_l]^T$. For a detailed summary of the DH convention, the reader is referred to~\cite{hartenberg1964kinematic}.


\section{Problem Formulation}
\label{sec:problem_formulation}

In this section, we present the calibration of a dynamic multi-camera cluster with one static camera, $s$, and one \emph{dynamic} camera, $d$, mounted to an actuated mechanism with L joints. We use a static fiducial marker of known size and assign to it a coordinate frame $\fr{\target}$. We first summarize the authors' previous DCC calibration approach~\cite{das2016calibration} which incorporates encoder measurements, and then describe our proposed method which performs the calibration of the DCC without the use of encoder feedback. Finally, we illustrate how a VIO algorithm can be adapted to simultaneously perform online estimation of the unknown mechanism joint angles and the VIO localization state.

\subsection{DCC Calibration With Encoder Feedback}
\label{subsec:cal_form}
\begin{figure}[tb]
	\centering
	\includegraphics[width=0.9\linewidth]{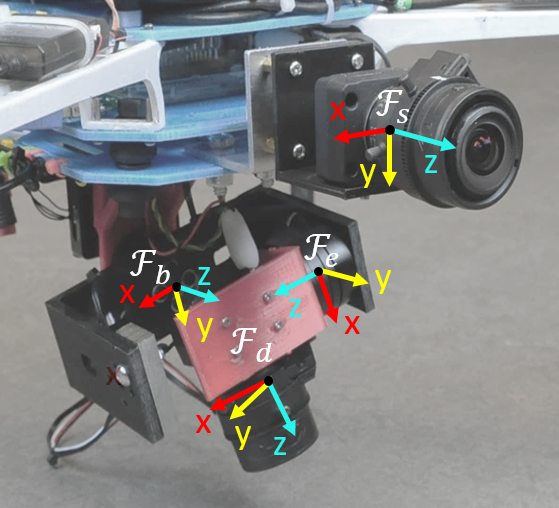}
	\caption{Frame diagram of a 2-DOF gimbal based DCC. Frames $\fr{s}$, $\fr{d}$, $\fr{b}$, $\fr{e}$ represent the static camera, dynamic camera, mechanism base, and mechanism end-effector frames respectively. An IMU is mounted to the back of the dynamic camera which is used for image stabilization of the gimbal and for estimating approximate joint angles to initialize the encoderless calibration.}
	\label{fig:2dof_gimbal}
\end{figure}

The aim of the calibration process is to determine the rigid body transformation $\Tpaf{\param}{\inp}{\dycam}{\stcam}$ from the static camera frame, $ \fr{\stcam} $, to the dynamic camera frame, $ \fr{\dycam} $, where $ \param $ is the set of estimated \emph{kinematic} parameters used to build the rigid body transform and $ \inp \in \mathbb{R}^L $, defined as $\inp = [\theta_1, \ldots,\theta_L]$, is the set of \emph{measured} parameters available from either known inputs to the mechanism, or measured encoder feedback. For the calibration, the transformation between cameras has the form $\Tpaf{\param}{\inp}{\dycam}{\stcam} = \Tpf{\sdof_\dycam}{\dycam}{\ee} \Tpaf{\dhp}{\inp}{\ee}{b} \Tpf{\sdof_\stcam}{b}{\stcam}$, where $\Tpf{\sdof_\stcam}{b}{\stcam}$ defines the transformation from the static camera to the mechanism base frame, $\Tpaf{\dhp}{\inp}{\ee}{b}$ defines the transformation from the base frame of the mechanism to the end effector frame, and $\Tpf{\sdof_\dycam}{\dycam}{\ee}$ defines the transformation from the end effector frame to the dynamic camera frame. Note that $\Tpaf{\dhp}{\inp}{\ee}{b}$ is a chain of transforms through the mechanism's links computed using its forward kinematics, and is a function of its DH parameters and control inputs. A DCC consisting of a 2-DOF gimbal is depicted in \Cref{fig:2dof_gimbal}.

For each instance where both the static and dynamic cameras capture measurements from the fiducial target, the pose of each camera with respect to the marker frame, $ \Tf{c}{t} $, $c \in \left\{s,d\right\}$, is determined by solving the Perspective-n-Point (PnP) problem~\cite{fischler1981random}. For an L joint mechanism, we define the $i$\textsuperscript{th} \emph{measurement set} as $ Z_i = \{ P_i^\stcam, P_i^\dycam, Q_i^\stcam, Q_i^\dycam, \inp_i \} $, where $ P_i^\stcam $, $ P_i^\dycam $ $\in \mathbb{R}^3$ are the set of corresponding marker point positions defined in the static and dynamic camera frames respectively, and are easily computed by passing the known target points through the target to camera transformation $ \Tf{c}{t}$. The set of pixel measurements for the target points, as observed by the static and dynamic cameras, are denoted by  $ Q_i^\stcam $ and $ Q_i^\dycam $ $\in \mathbb{R}^2$, respectively, and $ \inp_i$ is the set of known joint angles for the mechanism at snapshot $ i $.

Using the measurement set $i$ and the transformation between camera frames, we can now define the reprojection error between the measured marker point $j$ in the static camera frame and the corresponding measured point in the dynamic camera frame as

\begin{equation}
\label{eqn:err}
e_j^\dycam(\param, \inp_i) = z_j^\dycam - \proj^\dycam (\Tpaf{\param}{\inp_i}{\dycam}{\stcam} \point_j^\stcam)
\end{equation}
where $ z_j^\dycam \in Q_i^\dycam $ is the measurement of point $ j $, observed in the dynamic camera, and $ \point_j^\stcam \in P_i^\stcam $ is the 3D position of point j as observed from the static camera.  Since both the actuated and static camera observe the same marker at each snapshot, we can similarly compute the error for points observed in the actuated frame and projected into the static frame as 
\begin{equation}
e_j^\stcam(\param, \inp_i) = z_j^\stcam - \proj^\stcam ( (\Tpaf{\param}{\inp_i}{\dycam}{\stcam})^{-1} \point_j^\dycam)
\end{equation}
where $ z_j^\stcam \in Q_i^\stcam $ is the measurement of point $ j $ observed in the static camera, and $ \point_j^\dycam \in P_i^\dycam $  is the 3D position of point j as observed from the dynamic camera.  The total squared reprojection error as a function of the estimation parameters, $ \Lambda(\param): \mathbb{R}^n \mapsto \mathbb{R} $ over all of the collected measurement sets, $ \Gamma = \{Z_1, Z_2, \ldots, Z_k \} $ , is defined as
\begin{equation}
\label{eqn:cost}
\begin{split}
\Lambda(\param) = \sum\limits_{Z_i \in \Gamma} \sum\limits_{j=1}^{\card{P_i^\stcam}}  &e_j^\dycam(\param, \inp_i)^T e_j^\dycam(\param, \inp_i) \\
& + e_j^\stcam(\param, \inp_i)^T e_j^\stcam(\param, \inp_i)
\end{split}
\end{equation} 

Finally, an unconstrained optimization of~\eqref{eqn:cost} is performed in order to find the optimal calibration parameters, $ \param^* $, which minimize the total reprojection error over the set of collected measurements,

\begin{equation}
\label{eqn:opt} 
\param^* = \underset{\param}{\operatorname{argmin}} \, \Lambda(\param).
\end{equation}

Note that in~\eqref{eqn:opt}, the kinematic parameter vector, $ \param^* $, consists of the parameters required to define the transformation from the static camera frame to the base frame of the manipulator, the DH parameters of the actuated mechanism, and finally the parameters corresponding to the transformation from the end effector frame to the moving camera frame. The calibration process can also be formulated using a constrained optimization technique, and can be extended to perform a DCC calibration between any combination of static and dynamic camera pairs~\cite{das2016calibration}.

\subsection{DCC Calibration Without Encoder Feedback}
\label{sec:encoderlesscal}
In order to calibrate the DCC without the use of encoder feedback, the angles are added to the optimization as part of the estimated parameters. As a result, the size of the estimation parameters increases by $L$ with each new measurement. In addition to estimating the kinematic parameters, $\param$, we now simultaneously estimate the joint angles for each measurement configuration. We denote the estimated joint angles for the $i^{th}$ measurement set as $\beta_i = [\theta_1^i ~\cdots~ \theta_L^i]$. For the $i^{th}$ measurement set, we can write the full transformation from the static to the dynamic camera as 
\begin{equation}
\label{eqn:full_transform}
\Tpaf{\param}{\beta_i}{\dycam}{\stcam} = \Tpf{\sdof_\dycam}{\dycam}{\ee} \Tpaf{\dhp}{\beta_i}{\ee}{b} \Tpf{\sdof_\stcam}{b}{\stcam}
\end{equation}

The estimated joint angles for the mechanism, $\beta_i$ are required in order to determine the full transformation between cameras. Given a total of $K$ measurement sets, let us define the complete set of estimated joint angles as $\zeta = [\beta_1 ~ \cdots ~ \beta_K]$. The total squared reprojection error over the entire measurement set is then calculated as
\begin{equation}
\label{eqn:encoderless_cost}
\begin{split}
\Lambda(\param, \zeta) = \sum\limits_{Z_i \in \Gamma} \sum\limits_{j=1}^{\card{P_i^\stcam}}  &e_j^\dycam(\param, \beta_i)^T e_j^\dycam(\param, \beta_i) \\
& + e_j^\stcam(\param, \beta_i)^T e_j^\stcam(\param, \beta_i)
\end{split}
\end{equation} 

An unconstrained optimization is then performed on~\eqref{eqn:encoderless_cost} to determine the optimal kinematic calibration parameters, $\param^*$, and optimal joint angle values, $\zeta^*$, which minimize the total reprojection error,

\begin{equation}
\label{eqn:encoderlessopt} 
\param^*,\zeta^* = \underset{\param,\zeta}{\operatorname{argmin}} \, \Lambda(\param,\zeta)
\end{equation}

As a result of optimizing over both the kinematic parameters and the joint angles from each of the $K$ measurement set configurations, a DCC calibration with $L$ joints will have $12 + (3+K)L$ estimation parameters as opposed to $12 + 3L$ estimation parameters for the calibration case with known encoder angles~\cite{das2016calibration}. In order to prevent~\eqref{eqn:encoderlessopt} from converging to local minima, sufficiently accurate initialization values for the estimated kinematic and joint angle parameters are required. In practice, we have found that the kinematic parameters can be sufficiently initialized using approximate, hand-measured values, and the joint angles can be sufficiently initialized using approximate angles obtained from an IMU, as demonstrated in Section \ref{sec:real_calibration}.  Finally, it should be noted that while the joint angle values for the entire measurement set are estimated during the calibration of the DCC, the optimized joint angle values are not explicitly used afterwards when performing VIO or SLAM, and must be computed online as part of the VIO localization process.

\subsection{Visual Inertial Odometry using a DCC}

In order to perform VIO using a DCC, the unknown joint angles of the mechanism need to be estimated as the dynamic cluster is reconfigured. To that end, we extended OKVIS, the VIO implementation in~\cite{leutenegger2015keyframe}, to accommodate a DCC. OKVIS is an indirect, nonlinear optimization-based algorithm which is capable of online estimation of time-varying camera extrinsics by minimizing reprojection error. However, the original implementation models the extrinsics between cameras as a general transformation, and considers only small changes modeled by a Gaussian process, such as thermal expansion. Our extension incorporates the gimballed extrinsics model of~\eqref{eqn:full_transform}, and is capable of estimating joint angles with quick and unmodelled mechanism motion. 

We now briefly explain the approach presented in~\cite{leutenegger2015keyframe}.  Let us define the IMU and world frames as $\fr{I}$ and $\fr{W}$, respectively. The VIO state vector for the robot, $\mathbf{x}$, is estimated at every time-step $k$ of the algorithm, and is given by

\begin{equation}
\mathbf{x}^k = [ \mathbf{r}_{I}^{W} ~ \mathbf{q}^{W:I} ~ \mathbf{v}^{I} ~ \mathbf{b}_g ~ \mathbf{b}_a]^T
\end{equation}
where $\mathbf{r}_{I}^{W}$ denotes the position vector of the IMU frame of the cluster with respect to the world frame, $\mathbf{q}^{W:I}$ is the quaternion which defines the rotation between the IMU and world frame, $\mathbf{v}^{I}$ is the velocity of the IMU expressed in the IMU frame, and $\mathbf{b}_g $ and $\mathbf{b}_a $ are the bias states of the IMU's gyroscope and accelerometer, respectively. Let $\Tf{I}{W} \in \mathbb{SE}(3)$ be the rigid body transformation matrix composed using $ \mathbf{r}_{I}^{W}$ and $\mathbf{q}^{W:I} $, and $p_j^{W}$ be a 3D landmark point, estimated through keyframe triangulation.  Then, the point reprojection error term for the $i^{th}$ cluster camera is defined as

\begin{equation}
\label{eqn:okvis_reprojection}
e_j^i(\Tf{I}{W}) = z_j^i- \proj^i (\Tf{C_i}{I} \Tf{I}{W}p_j^{W})
\end{equation}
where $z_j^i$ is the pixel measurement in camera $i$ of landmark $p_j^{W}$, and $ \Tf{C_i}{I}$ is the extrinsic calibration between the IMU sensor and camera $C_i$.  The reprojection error term from~\eqref{eqn:okvis_reprojection} is combined with additional IMU and keyframe error terms, in order to optimize both the robot and landmark states.

In order to perform estimation of the mechanism joint angles, we augment the localization state vector with joint angles to be estimated at time $k$
\begin{equation}
\mathbf{x}^k_{\beta_k} = [ \mathbf{x}^k ~ \beta_k]^T
\end{equation}
where $ \beta_k = [\theta_1^k ~ \cdots ~ \theta_L^k]$ are the $L$ joint angles of the mechanism to be optimized.  When using a DCC, the extrinsics between the \emph{dynamic} camera and IMU, $\Tf{\dycam}{I}$, can be decomposed as

\begin{equation}
\label{eqn:dynamic_ext}
\Tpf{\beta_k}{\dycam}{I} = \Tpf{\sdof_\dycam}{\dycam}{\ee} \Tpaf{\dhp}{\beta_k}{\ee}{b} \Tpf{\sdof_\stcam}{b}{\stcam} \Tf{\stcam}{I}
\end{equation}
where $\Tf{\stcam}{I}$ is the transformation from IMU to static camera, which can be computed offline~\cite{furgale2013unified}. Using~\eqref{eqn:dynamic_ext}, the reprojection error for the dynamic camera can be written as
\begin{equation}
\label{eqn:okvis_reprojection_gimbal}
e_j^i(\Tf{I}{W},\beta_k) = z_j^i- \proj^i (\Tpf{\beta_k}{\dycam}{I}\Tf{I}{W}p_j^{W})
\end{equation}

Our extended implementation of OKVIS uses the modified reprojection error term from~\eqref{eqn:okvis_reprojection_gimbal}, along with the original IMU and keyframe error terms from the original implementation to estimate the augmented robot state vector $\mathbf{x}^k_{\beta_k} $, which includes the mechanism joint angles.  To perform the optimization, we analytically compute the Jacobian

\begin{equation}
\label{eqn:jac}
\frac{\partial  e_j^i(\Tf{I}{W},\beta_k)}{\partial \beta_k}
\end{equation}
which describes how the modified reprojection error from \Cref{eqn:okvis_reprojection_gimbal} is affected by small changes in the joint angles,~$\beta_k$.

\section{Experimental Validation}
\label{sec:experimental_validation}

To validate our proposed approach, we perform two sets of experiments.  In the first set, we demonstrate the successful encoderless calibration of a 2-DOF gimbal both in simulation and on physical hardware. In the second set of experiments, we perform visual inertial odometry using the calibrated gimbal, and show that our gimballed DCC VIO configuration performs comparably to a standard SCC setup.

\subsection{Encoderless Gimbal DCC Calibration}
 \textbf{Simulation:}  The encoderless calibration approach outlined in Section \ref{sec:encoderlesscal} is first validated in simulation. We generate a DCC consisting of a 2-DOF gimbal and collect synthetic measurements of a 9x7 chessboard fiducial target. Zero-mean error with 0.4 pixel standard deviation is added to the pixel measurements to simulate real-world image noise. To perform the calibration, we collect a calibration set of 81 independent image pairs from a wide range of gimbal configurations. In order to ensure sufficient configuration excitation for the calibration, we collect measurement sets using approximate uniform sampling of the 2-DOF joint angle space. An approach similar to~\cite{rebello2017autonomous} would optimize the required number of images to achieve adequate coverage of the configuration space. Prior to optimization, we initialize the kinematic and joint angle estimates with noisy measurements of the true value; this error is characterized as zero-mean noise with a standard deviation of 0.03~m and 10 degrees for the translation and rotation parameters, respectively.  
 
In order to verify the calibration, we independently collect a \emph{validation} set of 81 image pairs where the measurements of the fiducial target are taken at random angles within the 2-DOF joint angle space. Once the calibration is completed using the calibration set, we verify the results by performing a process similar to the description in Section \ref{sec:encoderlesscal}, except the kinematic parameters are held static at their calibrated values and only the joint angles are estimated. Using the estimated joint angle values, the reprojection error over the entire validation set is then computed. \Cref{tbl:sim_calibration_results} presents the average reprojection error for the calibration and validation sets. Since the error for both cases is low, at approximately 0.39 pixels, it is evident that our encoderless calibration procedure is able to perform a high quality fitting of the kinematic and joint angle parameters to the image data.

For the simulation study, both the true kinematic parameters and true joint angle values for each measurement set configuration are known. Thus, we also evaluate the error between these estimated parameters and their true values. \Cref{tbl:joint_errors} presents the average estimation error for the two joints from the simulated 2-DOF gimbal, while \Cref{tbl:trans_rot_error} gives a summary of the average parameter error of the kinematic parameters. The errors are presented separately for the translation parameters of the calibration (the translation components from $\sdof_d$ and the DH $d$ and $a$ values for each link), and the rotation parameters of the calibration (the rotation components from $\sdof_d$ and the DH $\alpha$ value for each link). For both the kinematic and joint angle parameters, the resultant low estimation error demonstrates the ability of our proposed approach to estimate the required parameters with high accuracy.  For this simulated case, it is apparent that the estimated parameters are observable, however, a full observability and degeneracy analysis for the encoderless DCC calibration problem is left as an area of future work. 

	 	\begin{table} [tb]
        	\caption{Reprojection error statistics of the simulated gimbal}
	 		\label{tbl:sim_calibration_results}
	 		\centering
	 		\begin{tabularx}{\columnwidth}{ l Y Y }
	 			\toprule
	 			Dataset & Calibration & Validation \\
	 			\midrule
	 			{Number of images} & 81 & 81\\
	 			{Average reprojection error (pixels)} & 0.3858  & 0.3854 \\ 
	 			{Standard deviation (pixels)} & 0.0183  & 0.0172 \\ 
	 			\bottomrule
	 		\end{tabularx}
	 	\end{table}
    
    \begin{table} [tb]
    	\centering
        \caption{Estimation error statistics of joint 1 and joint 2 (roll and pitch axes) of the simulated gimbal}
        \label{tbl:joint_errors}
        \begin{tabularx}{\columnwidth}{l Y Y}
        	\toprule
	 			Dataset & Calibration & Validation \\
            \midrule
            {Joint 1 average error (rad)} & $5.89 \times 10^{-3}$ &  $5.83 \times 10^{-3}$ \\
            {Joint 1 standard deviation (rad)} & $0.80 \times 10^{-3}$ & $0.75 \times 10^{-3}$\\
			{Joint 2 average error (rad)} & $2.38 \times 10^{-3}$ & $2.52 \times 10^{-3}$\\ 
            {Joint 2 standard deviation (rad)} & $0.71 \times 10^{-3}$ & $0.67 \times 10^ {-3}$\\
            \bottomrule
		\end{tabularx}
     \end{table}
     
     \begin{table} [tb]
    	\centering
         \caption{ Translation and rotation parameter error statistics of the DCC calibration}
         \label{tbl:trans_rot_error}
	 \begin{tabularx}{\columnwidth}{ l Y Y }
       \toprule
	 			Dataset & Calibration \\
       \midrule
       {Translation Parameter Average Error (m)} & $1.73 \times 10^{-3}$\\
       {Translation Parameter Error Standard Deviation (m)} & $7.93 \times 10^{-4}$ \\ 
       {Rotation Parameter Average Error (rad)} & $1.21 \times 10^{-3}$ \\
       {Rotation Parameter Error Standard Deviation (rad)} & $1.83 \times 10^{-3}$ \\
       \bottomrule
	 \end{tabularx}
     \end{table}
     
\begin{figure}[tb]
	\centering
	\includegraphics[width=0.9\linewidth]{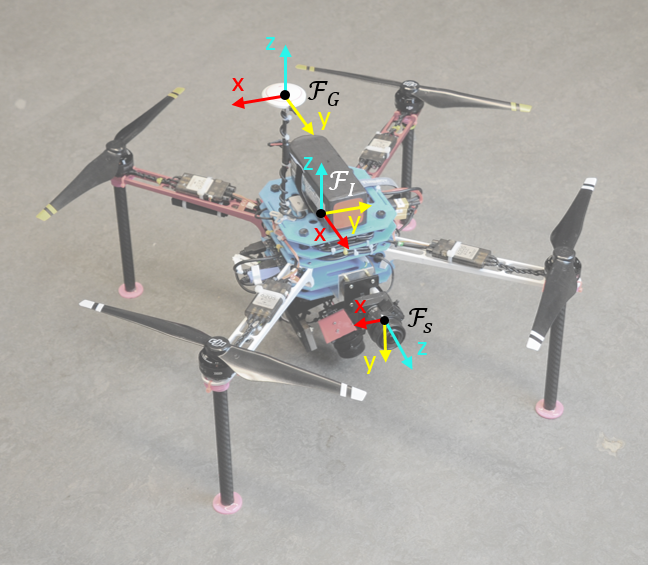}
	\caption{Custom quadrotor hardware with 2-DOF gimbal based DCC mounted on bottom. Frames $\fr{s}$, $\fr{I}$, $\fr{G}$ denote the static camera, IMU, and GPS frames, respectively.}
	\label{fig:quad}
\end{figure}

 \textbf{Physical Gimbal:} 
 \label{sec:real_calibration}
The proposed calibration approach is experimentally validated using a DCC consisting of a 2-DOF gimbal and two PointGrey Firefly MV cameras, which are software triggered at 15 fps and have a resolution of 640x480. The gimbal is controlled by an Alexmos SimpleBGC 32-bit gimbal motor controller, and the DCC is mounted onto a custom built quadrotor. The SCC and DCC configurations both have a baseline of approximately 10 cm. The 2-DOF gimbal, along with the labelled frames, is depicted in \Cref{fig:2dof_gimbal}, and the quadrotor is depicted in \Cref{fig:quad}.

Encoderless calibration of the gimballed DCC is performed using the method outlined in Section \ref{sec:encoderlesscal}.  To perform calibration, a calibration set of 83 independent image pairs is collected. Similar to the simulation study, the gimbal configurations are sampled uniformly over the 2-DOF joint angle space. An AprilGrid is used as the fiducial target for calibration, as accurate detection of this target is more robust to large viewpoint changes and partial target observations in comparison to a chessboard~\cite{furgale2013unified}. For validation of the gimbal calibration, an independent verification set of 70 image pairs is collected, also using the AprilGrid fiducial target. For both the calibration and validation set, we ensure that there is overlap of the fiducial target between the static and gimbal camera. Initialization of the joint angle values are provided by the gimbal IMU that is used for camera stabilization. Due to the gimbal geometry, the estimated roll and pitch values reported by the gimbal IMU roughly correspond to the gimbal joint angle rotation. The results for the calibration and verification sets are summarized in \Cref{tbl:real_calibration_results}.   

\begin{table} [tb]
    	\caption{Reprojection error statistics of the physical gimbal}
	   \label{tbl:real_calibration_results}
		\centering
	\begin{tabularx}{\columnwidth}{ l Y Y }
        \toprule
        Dataset & {Calibration} & {Validation}\\
        \midrule
        {Number of images} & 83 & 70\\
        {Average reprojection error (pixels)} & 1.5234  & 1.7933 \\ 
        {Standard deviation (pixels)} & 0.8487  & 1.1934 \\ 
        \bottomrule
	\end{tabularx}
\end{table}

We see that the average reprojection error for the calibration set is approximately 1.5 pixels, which indicates a good quality calibration was achieved.  The average error of the validation set is comparable to the result from the calibration set, at approximately 1.8 pixels, which corroborates the accuracy of the calibrated kinematic parameters. Note that this reprojection error is highly dependent on the quality of the intrinsic calibration, which can be reduced through the use of higher quality lenses. 

\subsection{VIO using Gimbal DCC}

\begin{figure}[tb]
  \centering
  \begin{subfigure}[t]{0.03\textwidth}
  a)
  \end{subfigure}
  \begin{subfigure}[t]{0.9\linewidth}  \hspace{-2em}
   \centering
  \includegraphics[width=\linewidth,valign=t]{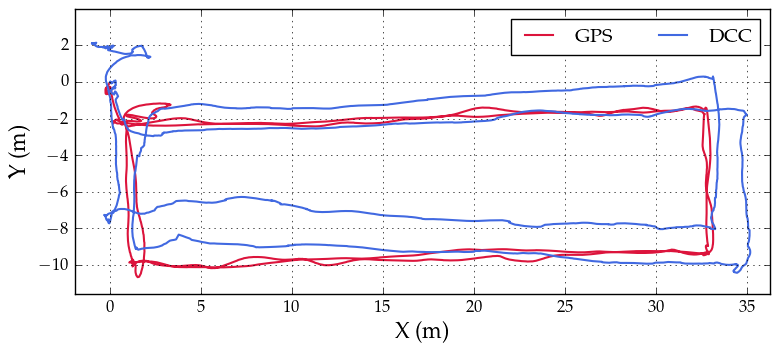}
  \end{subfigure}
  \begin{subfigure}[t]{0.03\textwidth}
  b)
  \end{subfigure}
  \begin{subfigure}[t]{0.9\linewidth}  \hspace{-2em}
  \centering
  \includegraphics[width=\linewidth,valign=t]{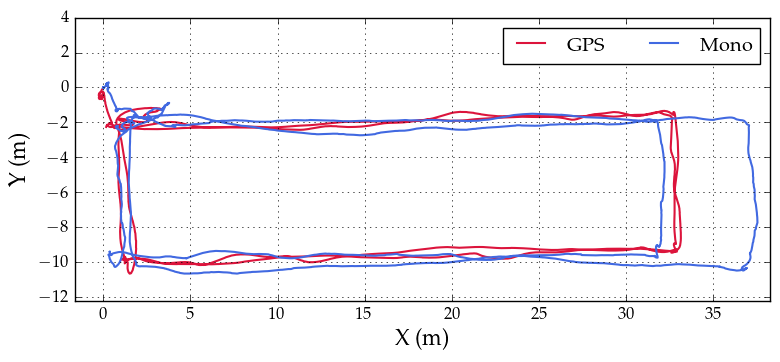}
  \end{subfigure}
  \begin{subfigure}[t]{0.03\textwidth}
  c)
  \end{subfigure}
  \begin{subfigure}[t]{0.9\linewidth}  \hspace{-2em}
  \centering
  \includegraphics[width=\linewidth,valign=t]{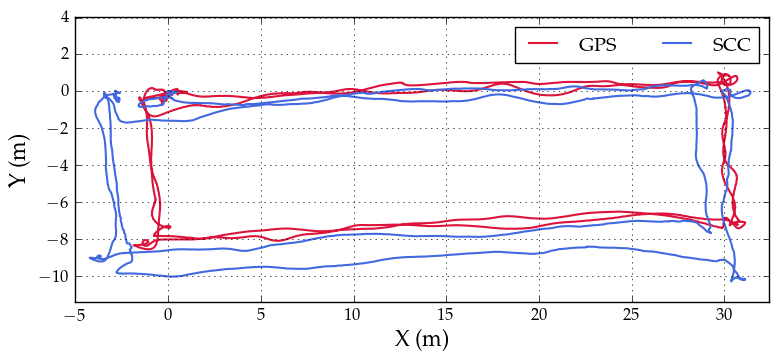}
  \end{subfigure}
  \caption{Estimated trajectories with (a) Dynamic Camera Cluster, (b) Monocular, and (c) Static Camera Cluster configurations. For all flight scenarios, the quadrotor starts at (0,0) and then flies two rectangular loops.}
  \label{fig:trajectories}
\end{figure}
\begin{figure}[tb]
  \centering
  \begin{subfigure}{0.48\linewidth}
    \centering
  	\includegraphics[width=0.90\linewidth]{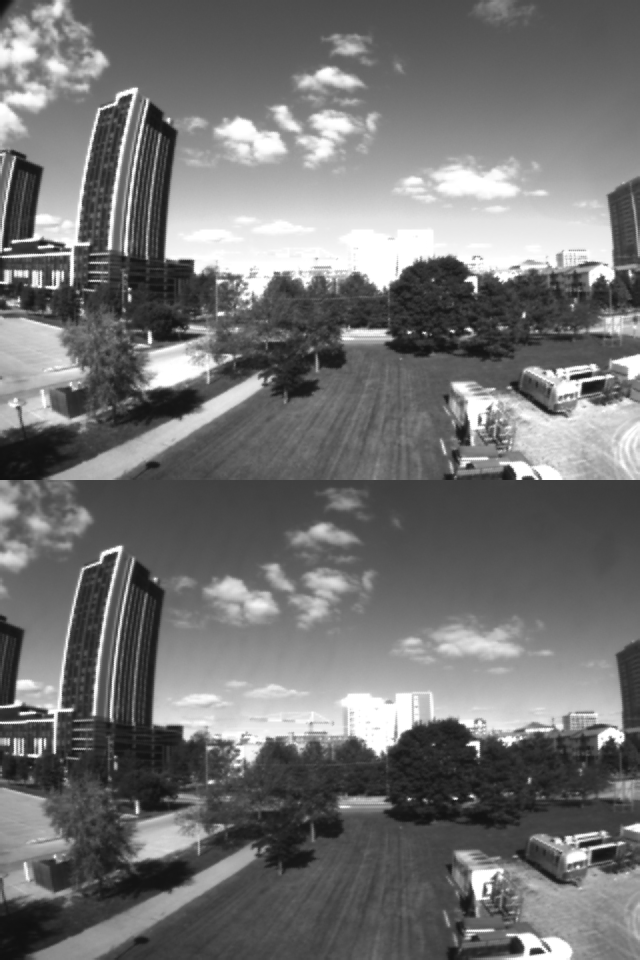}
    \caption{\label{fig:photo_static}}
  \end{subfigure}
  \begin{subfigure}{0.48\linewidth}
    \centering
	\includegraphics[width=0.90\linewidth]{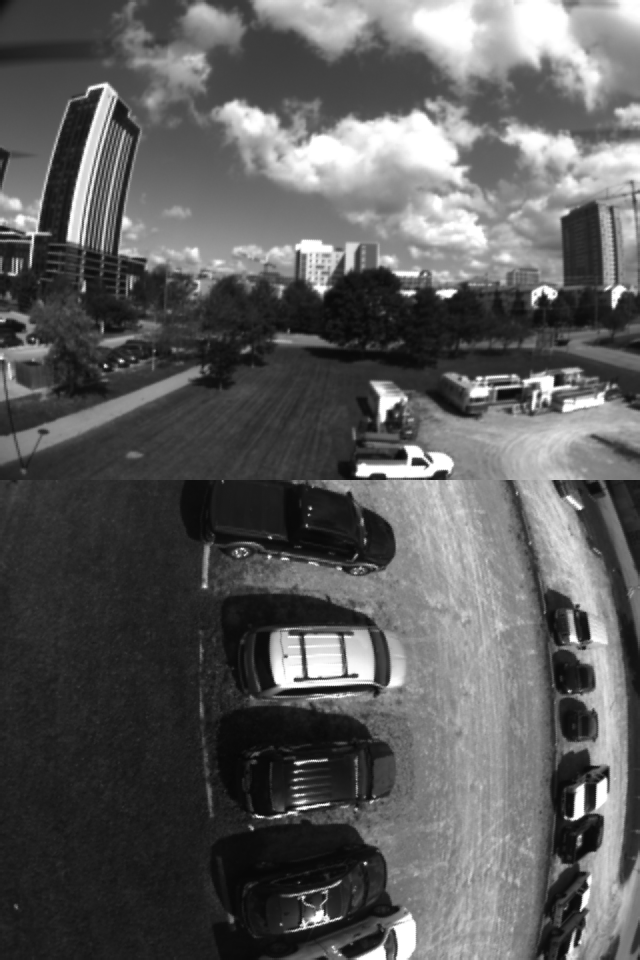}
	\caption{\label{fig:photo_gimbal}}
  \end{subfigure}
  \caption{Example frames from the two datasets. \subref{fig:photo_static} The static configuration, with one camera above the other. \subref{fig:photo_gimbal} The dynamic configuration, with the gimbal pointed down.}
  \label{fig:example_photos}
\end{figure}

\begin{table} [t]
	\centering
    \caption{Translation and rotation error of the three camera configurations, normalized by total distance travelled}
	\label{tbl:trajectory_error_comparison}
    \begin{tabularx}{\columnwidth}{p{2.7cm} Y Y Y}
      \toprule
      \quad & {DCC} & {Mono} & {SCC}\\
      \midrule
      Normalized translation \newline RMSE (\%) & $1.58 \times 10^{-2}$ & $1.24 \times 10^{-2}$ & $1.12 \times 10^{-2}$\\
      Normalized rotation \newline RMSE (rad/m) & $4.3 \times 10^{-4}$ & $2 \times 10^{-4}$ & $9.4 \times 10^{-4}$\\ 
      \bottomrule
    \end{tabularx}
\end{table}

To demonstrate our solution, we ran our extended OKVIS offline on a dataset collected using the quadrotor setup from \Cref{fig:quad}. The vehicle flew in the rectangular loop depicted in \Cref{fig:trajectories}, which is approximately 180~m in length. During the flight the gimbal DCC was pre-programmed to point towards features in the environment we deemed would benefit the VIO estimation. The effectiveness of gimbal viewpoint selection to improve pose estimation was not evaluated and is left as future work. We used Kalibr~\cite{furgale2013unified} to find the static transformation between the quadrotor IMU frame, $\fr{I}$, and the static camera frame, $\fr{s}$. The physical relationship between these frames is depicted in \Cref{fig:quad}. 

For validation, we compare three sets of results: (a) stereo VIO using the DCC shown in \Cref{fig:2dof_gimbal}, (b) monocular VIO using only the static camera from the DCC flight test, and (c) stereo VIO using a separate dataset with static cameras. These two flights followed approximately the same trajectory, and sample images from the two datasets are presented in \Cref{fig:example_photos}. Ground truth of the trajectory is collected using the DJI N3 Autopilot GPS, which has an accuracy of approximately 2~m standard deviation.  

The three tested configurations all expectedly experience trajectory drift compared to the GPS ground truth solution. We hypothesize the drift observed could be due to experimental factors, such as the high altitude of the flown trajectory and the lack of IMU excitation. The quadrotor's flight altitude of 10 m in conjunction with the small camera baseline results in inaccurate feature depth initialization and a high average feature distance, which in turn causes poor scale observability. In addition to this, the constant-velocity trajectory executed by the quadrotor inhibits IMU excitation, which could be another factor contributing to poor scale recovery. The relative performance of each configuration is consistent, as shown in Table \ref{tbl:trajectory_error_comparison}, which reports the RMSE of the trajectory normalized by the total distance flown.

The results suggest that the DCC result exhibits a slightly larger normalized translation error compared to the mono and SCC trials. This can be attributed to some of the required modifications to the OKVIS implementation. OKVIS includes a 3D-2D RANSAC step which selects landmark measurements based on their predicted motion, which relies on the camera extrinsics. In the case of the dynamic camera, however, the measurements filtered out by RANSAC are precisely those required to estimate the joint angles for the time-varying extrinsics. Therefore, we disabled the 3D-2D RANSAC step for the dynamic camera. A possible future refinement would be a two-stage approach: first optimizing without rejecting outliers to obtain the dynamic camera extrinsics, then re-optimizing with the updated extrinsics for an improved motion estimate. Despite the slightly larger error for the DCC trial, it is evident that all configurations performed comparably, which verifies that the integration of the DCC did not significantly change the localization performance of the algorithm.

\begin{figure}[htb]
  \centering
  \includegraphics[width=\linewidth]{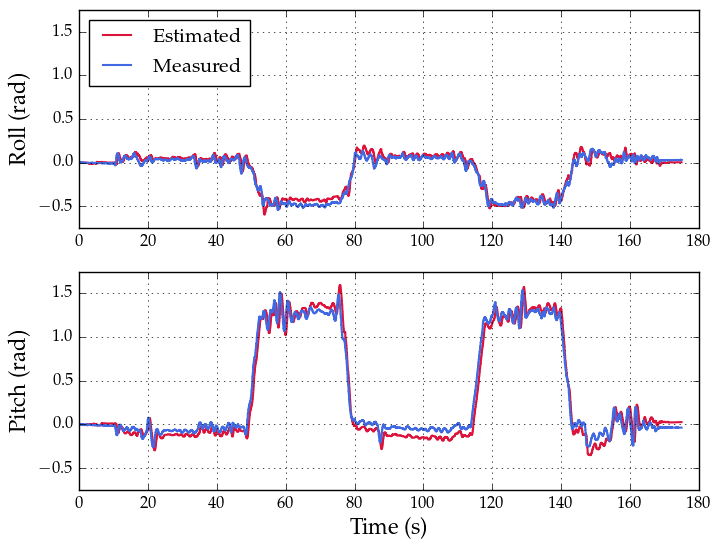}
  \caption{Estimated gimbal joint angles compared to ground truth provided by the gimbal encoders.}
  \label{fig:okvis_gimbal_angles}
\end{figure}

\Cref{fig:okvis_gimbal_angles} compares the estimated gimbal joint angles to the values obtained using encoders mounted on the gimbal for ground truth collection, and \Cref{tbl:gimbal_joint_angle_error} reports the RMSE of the estimated joint angles. We see that as the gimbal changes configuration, our extended VIO algorithm is able to accurately estimate the joint angles. It is important to state that no motion model or input from the gimbal IMU was used for the estimation, and only visual data from the camera images were used. A component of the errors reported in \Cref{tbl:gimbal_joint_angle_error} are due to a noticeable time lag in the joint angle estimation, which can be improved through hardware synchronization in the DCC. 
	 	        
\begin{table} [tb]
    \centering
    \caption{Roll and Pitch RMSE for the DCC configuration.}
	\label{tbl:gimbal_joint_angle_error}
    \begin{tabularx}{\columnwidth}{l Y}
        \toprule
        \quad & RMSE (rad) \\
        \midrule
        {Roll angle} & $4.5 \times 10^{-2}$\\
        {Pitch angle} & $9.2 \times 10^{-2}$\\
        \bottomrule
    \end{tabularx}
\end{table}

\section{Conclusion}
\label{sec:conclusion}

This work presents a method to perform encoderless calibration of a dynamic camera cluster. Our approach simultaneously estimates kinematic parameters of the DCC, as well as the joint angles of the mechanism for each measurement of a fiducial target. We demonstrate successful calibration of a 2-DOF gimbal DCC both in simulation and using physical hardware. The calibrated values are then used to perform VIO on a custom quadrotor, using an extended implementation of OKVIS which also performs online estimation of the gimbal joint angles. We demonstrate that the extended VIO is able to successfully estimate the joint angles, and that it performs comparably to a VIO solution using a standard static camera configuration. Future work will include testing on a wider range of actuated mechanisms, degeneracy analysis of both the encoderless calibration and online gimbal joint angle estimation, as well as incorporating the calibration into an \emph{active} Visual SLAM framework.

\bibliography{ref}{}

\begin{thebibliography}{10}

\bibitem{tribou2015multi}
M.~J. Tribou, A.~Harmat, D.~W. Wang, I.~Sharf, and S.~L. Waslander,
  ``Multi-camera parallel tracking and mapping with non-overlapping fields of
  view,'' {\em The International Journal of Robotics Research}, vol.~34,
  no.~12, pp.~1480--1500, 2015.

\bibitem{heng2015self}
L.~Heng, G.~H. Lee, and M.~Pollefeys, ``Self-calibration and visual slam with a
  multi-camera system on a micro aerial vehicle,'' {\em Autonomous Robots},
  vol.~39, no.~3, pp.~259--277, 2015.

\bibitem{manderson2016texture}
T.~Manderson, F.~Shkurti, and G.~Dudek, ``Texture-aware slam using stereo
  imagery and inertial information,'' in {\em Conference on Computer and Robot
  Vision (CRV)}, pp.~456--463, 2016.

\bibitem{frintrop2008attentional}
S.~Frintrop and P.~Jensfelt, ``Attentional landmarks and active gaze control
  for visual slam,'' {\em IEEE Transactions on Robotics}, vol.~24, no.~5,
  pp.~1054--1065, 2008.

\bibitem{das2016calibration}
A.~Das and S.~L. Waslander, ``Calibration of a dynamic camera cluster for
  multi-camera visual slam,'' in {\em IEEE/RSJ International Conference on
  Intelligent Robots and Systems (IROS)}, pp.~4637--4642, 2016.

\bibitem{rebello2017autonomous}
J.~Rebello, A.~Das, and S.~L. Waslander, ``Autonomous active calibration of a
  dynamic camera cluster using next-best-view,'' in {\em IEEE/RSJ International
  Conference on Intelligent Robots and Systems (IROS)}, 2017.

\bibitem{leutenegger2015keyframe}
S.~Leutenegger, S.~Lynen, M.~Bosse, R.~Siegwart, and P.~Furgale,
  ``Keyframe-based visual--inertial odometry using nonlinear optimization,''
  {\em The International Journal of Robotics Research}, vol.~34, no.~3,
  pp.~314--334, 2015.

\bibitem{horaud1995hand}
R.~Horaud and F.~Dornaika, ``Hand-eye calibration,'' {\em The International
  Journal of Robotics Research}, vol.~14, no.~3, pp.~195--210, 1995.

\bibitem{remy1997hand}
S.~Remy, M.~Dhome, J.-M. Lavest, and N.~Daucher, ``Hand-eye calibration,'' in
  {\em IEEE/RSJ International Conference on Intelligent Robots and Systems
  (IROS)}, vol.~2, pp.~1057--1065, 1997.

\bibitem{chen1993new}
C.~Chen and Y.~F. Zheng, ``A new robotic hand/eye calibration method by active
  viewing of a checkerboard pattern,'' in {\em IEEE International Conference on
  Robotics and Automation (ICRA)}, pp.~770--775, 1993.

\bibitem{kim2010robot}
S.-J. Kim, M.-H. Jeong, J.-J. Lee, J.-Y. Lee, K.-G. Kim, B.-J. You, and S.-R.
  Oh, ``Robot head-eye calibration using the minimum variance method,'' in {\em
  IEEE International Conference on Robotics and Biomimetics (ROBIO)},
  pp.~1446--1451, 2010.

\bibitem{pradeep2014calibrating}
V.~Pradeep, K.~Konolige, and E.~Berger, ``Calibrating a multi-arm multi-sensor
  robot: A bundle adjustment approach,'' in {\em Experimental Robotics},
  pp.~211--225, Springer, 2014.

\bibitem{hartenberg1964kinematic}
R.~S. Hartenberg and J.~Denavit, {\em Kinematic synthesis of linkages}.
\newblock McGraw-Hill, 1964.

\bibitem{kawamura2014encoderless}
A.~Kawamura, M.~Tachibana, S.~Yamate, and S.~Kawamura, ``Encoderless robot
  motion control using vision sensor and back electromotive force,'' in {\em
  IEEE/RSJ International Conference on Intelligent Robots and Systems (IROS)},
  pp.~1609--1615, 2014.

\bibitem{baik2004position}
W.-S. Baik, M.-H. Kim, N.-H. Kim, and D.-H. Kim, ``Position sensorless control
  system of srm using neural network,'' in {\em IEEE 35th Annual Power
  Electronics Specialists Conference}, vol.~5, pp.~3471--3475, 2004.

\bibitem{kormushev2015encoderless}
P.~Kormushev, Y.~Demiris, and D.~G. Caldwell, ``Encoderless position control of
  a two-link robot manipulator,'' in {\em IEEE International Conference on
  Robotics and Automation (ICRA)}, pp.~943--949, 2015.

\bibitem{li20133}
M.~Li and A.~I. Mourikis, ``3-d motion estimation and online temporal
  calibration for camera-imu systems,'' in {\em IEEE International Conference
  on Robotics and Automation (ICRA)}, pp.~5709--5716, 2013.

\bibitem{jia2013online}
C.~Jia and B.~L. Evans, ``Online calibration and synchronization of cellphone
  camera and gyroscope,'' in {\em IEEE Global Conference on Signal and
  Information Processing (GlobalSIP)}, pp.~731--734, 2013.

\bibitem{weiss2012real}
S.~Weiss, M.~W. Achtelik, S.~Lynen, M.~Chli, and R.~Siegwart, ``Real-time
  onboard visual-inertial state estimation and self-calibration of mavs in
  unknown environments,'' in {\em IEEE International Conference on Robotics and
  Automation (ICRA)}, pp.~957--964, 2012.

\bibitem{joukov2017generalized}
V.~Joukov, J.~F.-S. Lin, and D.~Kulic, ``Generalized hebbian algorithm for
  wearable sensor rotation estimation,'' in {\em IEEE/RSJ International
  Conference on Intelligent Robots and Systems (IROS)}, 2017.

\bibitem{kummerle2012simultaneous}
R.~K{\"u}mmerle, G.~Grisetti, and W.~Burgard, ``Simultaneous parameter
  calibration, localization, and mapping,'' {\em Advanced Robotics}, vol.~26,
  no.~17, pp.~2021--2041, 2012.

\bibitem{carrera2011slam}
G.~Carrera, A.~Angeli, and A.~J. Davison, ``Slam-based automatic extrinsic
  calibration of a multi-camera rig,'' in {\em IEEE International Conference on
  Robotics and Automation (ICRA)}, pp.~2652--2659, 2011.

\bibitem{warren2013online}
M.~Warren, D.~McKinnon, and B.~Upcroft, ``Online calibration of stereo rigs for
  long-term autonomy,'' in {\em IEEE International Conference on Robotics and
  Automation (ICRA)}, pp.~3692--3698, 2013.

\bibitem{bajcsy1988active}
R.~Bajcsy, ``Active perception,'' {\em Proceedings of the IEEE}, vol.~76,
  no.~8, pp.~966--1005, 1988.

\bibitem{sharp2001vision}
C.~S. Sharp, O.~Shakernia, and S.~S. Sastry, ``A vision system for landing an
  unmanned aerial vehicle,'' in {\em IEEE International Conference on Robotics
  and Automation (ICRA)}, vol.~2, pp.~1720--1727, 2001.

\bibitem{playle2015improving}
N.~Playle, ``Improving the performance of monocular visual simultaneous
  localisation and mapping through the use of a gimballed camera,'' Master's
  thesis, University of Toronto (Canada), 2015.

\bibitem{borowczyk2017autonomous}
A.~Borowczyk, D.-T. Nguyen, A.~Phu-Van~Nguyen, D.~Q. Nguyen, D.~Saussi{\'e},
  and J.~Le~Ny, ``Autonomous landing of a quadcopter on a high-speed ground
  vehicle,'' {\em Journal of Guidance, Control, and Dynamics}, 2017.

\bibitem{fischler1981random}
M.~A. Fischler and R.~C. Bolles, ``Random sample consensus: a paradigm for
  model fitting with applications to image analysis and automated
  cartography,'' {\em Communications of the ACM}, vol.~24, no.~6, pp.~381--395,
  1981.

\bibitem{furgale2013unified}
P.~Furgale, J.~Rehder, and R.~Siegwart, ``Unified temporal and spatial
  calibration for multi-sensor systems,'' in {\em IEEE/RSJ International
  Conference on Intelligent Robots and Systems (IROS)}, pp.~1280--1286, 2013.

\end{thebibliography}
\bibliographystyle{ieeetr}

\end{document}